\RequirePackage{fix-cm}
\documentclass[twocolumn]{svjour3}          
\smartqed  
\usepackage{graphicx}
\usepackage{url}
\usepackage{subfig}
\journalname{IJDL}
\begin{document}

\title{Identifying Reference Spans:\\ Topic Modeling and Word Embeddings help IR\thanks{Research supported in part by NSF grants CNS 1319212, DGE 1433817, and DUE 1241772}
}

\author{Luis Moraes \and 
	    Shahryar Baki \and
	    Rakesh Verma \and
	    Daniel Lee
}

\institute{L. Moraes \and S. Baki \and R. Verma \and D. Lee
            \at
		      Dept. of Computer Science, University of Houston, Houston, TX 77204 \\
              \email{ltdemoraes@outlook.com}             \\
		      \email{sh.baki@gmail.com}             \\
		      \email{rmverma@cs.uh.edu}             \\
		      \email{dlseven777@gmail.com}           
}

\date{Received: date / Accepted: date}

\maketitle

\begin{abstract}

The CL-SciSumm 2016 shared task introduced an interesting problem: given a document $D$ and a piece of text that cites $D$, how do we identify the text spans of $D$ being referenced by the piece of text? The shared task provided the first annotated dataset for studying this problem. We present an analysis of our continued work in improving our system's performance on this task. We demonstrate how topic models and word embeddings can be used to surpass the previously best performing system.

\end{abstract}

\section{Introduction}
\label{intro}

The CL-SciSumm 2016 \cite{jaidka2014computational} shared task posed the problem of automatic summarization in the computational linguistics domain. Single document summarization is hardly new ~\cite{RPL07,bvv11,rvcicling}; however, in addition to the reference document to be summarized, we are also given citances, i.e.\ sentences that cite our reference document. The usefulness of citances in the process of summarization is immediately apparent. A citance can hint at what is interesting about the document. 

This objective was split into three tasks. Given a citance (a sentence containing a citation), in Task 1a we must identify the span of text in the reference document that best reflects what has been cited. Task 1b asks us to classify the cited aspect according to a predefined set of facets: hypothesis, aim, method, results, and implication. Finally, Task 2 is the generation of a structured summary for the reference document. Although the shared task is broken up into multiple tasks, this paper concerns itself solely with Task 1a.

Task 1a is quite interesting all by itself. We can think of Task 1a as a small scale summarization. Thus, being precise is incredibly important: the system must often find a single sentence among hundreds (in some cases, however, multiple sentences are correct). The results of the workshop \cite{BIRNDL_PPT} reveal that Task 1a is quite challenging. There was a varied selection of methods used for this problem: SVMs, neural networks, learning-to-rank algorithms, and more.
Regardless, our previous system had the best performance on the test set for CL-SciSumm: cosine similarity between weighted bag-of-word vectors. The weighting used is well known in information retrieval: term frequency $\cdot$ inverse document frequency (TFIDF). Although TFIDF is a well known and understood method in information retrieval, it is surprising that it achieved better performance than more heavily engineered solutions. Thus, our goal in this paper is twofold: to analyze and improve on the performance of TFIDF and to push beyond its performance ceiling.

In the process of exploring different configurations, we have observed the performance of our TFIDF method vary substantially. Text preprocessing parameters can have a significant effect on the final performance. This variance also underscores the need to start with a basic system and then add complexity step-by-step in a reasoned manner.
Another prior attempt employed SVMs with tree kernels but the performance never surpassed that of TFIDF. Therefore, we focus on improving the TFIDF approach.

Depending on the domain of your data, it can be necessary to start with simple models. 
In general, unbalanced classification tasks are hard to evaluate due to the performance of the baseline. For an example that is not a classification task, look no further than news articles: the first few sentences of a news article form an incredibly effective baseline for summaries of the whole article.

First, we study a few of the characteristics of the dataset.
In particular, we look at the sparsity between reference sentences and citances, what are some of the hurdles in handling citances, and whether chosen reference sentences appear more frequently in a particular section. 
Then we cover improvements to TFIDF. We also introduce topic models learned through Latent Dirichlet Allocation (LDA) and word embeddings learned through word2vec. These systems are studied for their ability to augment our TFIDF system. Finally, we present an analysis of how humans perform at this task.

\subsection{Related Work}
\label{sec:rel_work}
Past research has already shown the importance of citations as a source of salient information for extractive summarization. 
There is a lot of work in trying to summarize a scientific paper using the articles that cite it.

The huge influx of academic research and work is not a new phenomenon. As databases of information grow, so too does the need to quickly sift through and find the important ``needles'' in the proverbial haystack. This need has brought continued attention to the area of summarization. One of the results of this focus was the CL-SciSumm 2016 shared task \cite{jaidka2014computational}.

In the pilot task, we focus on citations and the text spans they cite in the original article. The importance of citations for summarization is discussed in \cite{qazvinian2013generating}, which compared summaries that were based on three different things: only the reference article; only the abstract; and, only citations. The best results were based on citations. Mohammad et al. \cite{mohammad2009using} also showed that the information from citations is different from that which can be gleaned from just the abstract or reference article. However, it is cautioned that citations often focus on very specific aspects of a paper \cite{elkiss2008blind}.

Because of this recognized importance of citation information, research has also been done on properly tagging or marking the actual citation. Powley and Dale \cite{powley2007evidence} give insight into recognizing text that is a citation. Siddharthan and Teufel demonstrate how this is useful in reducing the noise when comparing citation text to reference text \cite{siddharthan2007whose}. Siddharthan and Teufel also introduce ``scientific attribution'' which can help in discourse classification. The importance of discourse classification is further developed in \cite{abu2011coherent}: they were able to show how identifying the discourse facets helps produce coherent summaries.

The choice of proper features is very important in handling citation text. Previous research \cite{kupiec1995trainable,besagni2003segmentation} gives insight into these features. We find in \cite{kupiec1995trainable} an in-depth analysis of the usefulness of certain features. As a result, we have used it to guide our selection of which features to include.

In addition to these features, we have to consider that multiple citation markers may be present in a sentence. Thus, only certain parts of a sentence may be relevant to identifying the target of a particular citation marker. Qazvinian and Radev \cite{qazvinian2010identifying} share an approach to find the fragment of a sentence that applies to a citation, especially in the case of sentences with multiple citation markers. The research of Abu-Jbara and Radev \cite{abu2012reference} further argues that a fragment need not always be continguous.

\subsection{CL-SciSumm 2016}

We present a short overview of the different approaches used to solve Task 1a.

Aggarwal and Sharma \cite{as16} use bag-of-words bigrams, syntactic dependency
cues and a set of rules for extracting parts of referenced documents
that are relevant to citances. 

In \cite{krk16}, researchers generate three combinations of an unsupervised graph-based
sentence ranking approach with a supervised classification approach. In 
the first approach, sentence ranking is modified to use information provided
by citing documents. In the second, the ranking procedure is applied as a
filter before supervised classification. In the third, supervised
learning is used as a filter to the cited document, before sentence
ranking.

Cao et al. \cite{clw16} model Task 1a as a ranking problem and apply SVM
Rank for this purpose. 

In \cite{lmlx16}, the citance is treated as a query over the sentences
of the reference document. They used learning-to-rank algorithms
(RankBoost, RankNet, AdaRank, and Coordinate Ascent) for
this problem with lexical (bag-of-words features), topic features and
TextRank for ranking sentences. WordNet is used to compute concept
similarity between citation contexts and candidate spans. 

Lei et al. \cite{lmz16} use SVMs and rule-based methods with lexicon features
(high frequency words within the reference text, LDA to train the reference document
and citing documents, and co-occurrence lexicon) and
similarities (IDF, Jaccard, and context similarity). 

In \cite{nomoto16}, authors propose a linear combination between a TFIDF model and a single layer
neural network model. This paper is the most similar to our work.

Saggion et al. \cite{sar16} use supervised algorithms with feature
vectors representing the citance and reference document
sentences. Features include positional, WordNet similarity measures,
and rhetorical features. 

We have chosen to use topic modeling and word embeddings to overcome the weaknesses of the TFIDF approach. Another participant of the CL-SciSumm 2016 shared task did the same \cite{lmz16}. Their system performed well on the development set, but not as well on the held-out test set. We show how improving a system with a topic model or a word embedding is a lot less straightforward than expected.

\section{Preliminaries}
Following are brief explanations of terms that will be used throughout the paper.

\textbf{Cosine Similarity.} This is a measure of similarity between two non zero vectors, $A$ and $B$ that measure the cosine of angle between them. Equation \ref{eq1} shows the formula for calculating cosine similarity.
\begin{equation}
  similarity(A, B) = \cos{\theta} = \frac{A \cdot B}{\left \| A \right \|\left \| B \right \|} \label{eq1}
\end{equation}

In the above formula, $\theta$ is the angle between the two vectors $A$ and $B$. We use cosine similarity to measure how far or close two sentences are from each other and rank them based on their similarity. In our task, each vector represents TFIDF or LDA values for all the words in a sentence. The higher the value of \\ $similarity(A,B)$, the greater the similarity is between the two sentences. 

\textbf{TFIDF.} This is short for \textit{term frequency-inverse document frequency}, and is a common scoring metric used for words in a query across a corpus of documents. The metric tries to capture the importance of a word by valuing frequency of the words use in a document and devaluing its appearance in every document. This was originally a method for retrieving documents from a corpus (instead of sentences from a document). For our task of summarization, this scoring metric was adjusted to help select matching sentences, so each sentence is treated as a document for our purposes. Thus, our ``document'' level frequencies are the frequencies of words in a sentence. The ``corpus'' will be the whole reference document. Then, the term frequency can be calculated by counting a word's frequency within a sentence. The inverse document frequency of a word will be based on the number of sentences that contain that word. When using TFIDF for calculating similarity, we use Equation~\ref{eq1} where the vectors are defined as:
\begin{equation}
  A = \langle a_1, a_2, \ldots, a_n \rangle \quad\quad \texttt{where} \quad\quad a_i = tf_{w_i} \cdot idf_{w_i} \label{eq:vec-tfidf}
\end{equation}

\begin{equation}
  idf_{w_i} = \log ( N / df_{w_i} )
\end{equation}
where $tf_{w_i}$ refers to the term frequency of $w_i$, $df_{w_i}$ refers to the document frequency of $w_i$ (number of documents in which $w_i$ appears), and $N$ refers to the total number of documents. 

\textbf{WordNet.} This is a large lexical dataset for the English language \cite{miller1995wordnet}. The main relation among words in WordNet is synonymy. However, it contains other relations like antonymy, hyperonymy, hyponymy, meronymy, etc. For our task of summarization, we use synonymy for expanding words in reference sentences and citances. Since reference sentences and citances are written by two different authors, adding synonyms increases the chance of a word occurring in both sentences if they are both indeed related.

\textbf{LDA.} Latent Dirichlet Allocation is a technique used for topic modeling. It learns a generative model of a document. Topics are assumed to have some prior distribution, normally a symmetric Dirichlet distribution. Terms in the corpus are assumed to have a multinomial distribution. These assumptions form the basis of the method. After learning the parameters from a corpus, each term will have a topic distribution which can be used to determine the topics of a document. When using LDA for calculating similarity, we use Equation~\ref{eq1} where the vectors are defined as topic membership probabilities:
\begin{equation}
  A = \langle a_1, a_2, \ldots, a_n \rangle \quad \texttt{where} \quad a_i = P( doc_A \in \, topic_i) \label{eq:vec-lda}
\end{equation}

\textbf{$F_1$-score.} To evaluate our methods, we have chosen the $F_1$-score. The $F_1$-score is a weighted average of precision and recall, where precision and recall receive equal weighting. This kind of weighted average is also referred to as the harmonic mean. Precision is the proportion of correct results among the results that were returned. And recall is the proportion of correct results among all possible correct results.

Our system outputs the top 3 sentences and we compute recall, precision, and $F_1$-score using these sentences. If a relevant sentence appears in the top 3, then it factors into recall, precision, and $F_1$-score. Thus, we naturally present 
the \emph{precision at $N$ measure} ($P@N$) used by \cite{campos2015evaluation}. Precision at $N$ is simply the proportion of correct results in the top $N$ ranks. In our evaluations, $N=3$. Average precision and the area under the ROC curve are two other measures that present a more complete picture when there is a large imbalance between classes. To keep in line with the evaluation for the BIRNDL shared task we chose to use P@N.
Regardless, we focus on the $F_1$-score rather than $P@3$ when determining if one system is better than another. 

If we look at the percentage of sentences that appear in the gold standard, we see that roughly 90\% of the sentences in our dataset are never chosen by an annotator. This means our desired class is rare and diverse, similar to outliers or anomalies \cite{campos2015evaluation}. Therefore, we should expect low performance from our system since our task is similar to anomaly detection \cite{campos2015evaluation} which has a hard time achieving good performance in such cases.

\section{Dataset}
\label{sec:dataset}

The dataset~\cite{cl-scisumm} consists of 30 total documents separated into three sets of 10 documents each: training, development, and test sets. For the following analysis, no preprocessing has been done (for instance, stemming).

There are 23356 unique words among the reference documents in the dataset. The citances contain 5520 unique words. The most frequent word among reference documents appears in 4120 sentences. The most frequent word among citances appears in 521 sentences. There are 6700 reference sentences and 704 citances (although a few of these should actually be broken up into multiple sentences). The average reference sentence has approximately 22 words in this dataset whereas citances have an average of approximately 34 words.

\begin{figure}[t]
	\centering
	\includegraphics[width=0.4\textwidth]{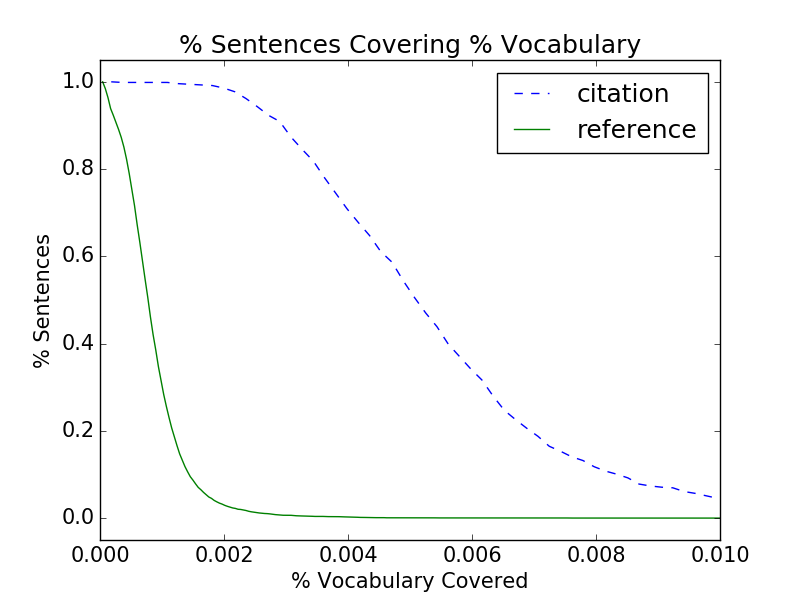}
	\caption{The percentage of sentences that contain a percentage of all unique words. An indirect measure of sparsity.}
	\label{fig:decay}
\end{figure}

In Figure~\ref{fig:decay} we can see the sparsity of the dataset. At a particular $(x,y)$ along the curves we know $x\%$ of all sentences contain at least some number of unique words -- a number equal to $y\%$ of the vocabulary. All sentences contain at least one word, which is a very small sliver of the vocabulary (appearing as 0\% in the graph). The quicker the decay, the greater the sparsity. Noise in the dataset is one of the factors for the sparsity. We can see that citances, seen as a corpus, are in general less sparse than the reference texts. This can be an indication that citances have some common structure or semantics.

One possibility is that citance must have some level of summarization ability. If we look at the annotations of a document as a whole we see a pattern: the annotators tend to choose from a small pool of reference sentences. Therefore, the sentences chosen are usually somewhat general and serve as tiny summaries of a single concept. Furthermore, the chosen reference sentences make up roughly 10\% of all reference sentences from which we have to choose.

\subsection{Citances}

It should be noted that citances have a few peculiarities, such as an abundance of citation markers and proper names. Citation markers (cues in written text demarcating a citation) will sometimes include the names of authors, thus the vocabulary for these sentences will include more proper names. This could justify the lesser sparsity if authors reoccur across citances. However, it could also justify greater sparsity since these authors may be unique. Identifying and ignoring citation markers should reduce noise. A preprocessing step we employ with this goal is the removal of all text enclosed in brackets of any kind. 

To demonstrate the differences in difficulty a citance can pose we present two examples: one that is relatively simple and another that is relatively hard. In both examples the original citance marker is in \textit{italics}.

\begin{quotation}

\textbf{Easy Citance}: ``According to \textit{Sproat et al. (1996)}, most prior work in Chinese segmentation has exploited \textbf{lexical} knowledge bases; indeed, the authors assert that they were aware of only one previously \textbf{published instance} (the mutual-information method of \textbf{Sproat and Shih} (1990)) of a purely \textbf{statistical approach}.''

\textbf{Reference Span}: ``Roughly speaking, previous work can be divided into three categories, namely purely \textbf{statistical} \textbf{approaches}, purely \textbf{lexical} rule-based approaches, and approaches that combine \textbf{lexical} information with statistical information. The present proposal falls into the last group. Purely \textbf{statistical} \textbf{approaches} have not been very popular, and so far as we are aware earlier work by \textbf{Sproat and Shih} (1990) is the only \textbf{published instance} of such an approach.''
\end{quotation}

In the ``Easy'' case, there are many salient words in common between the reference spans we must retrieve and the citance. This is the ideal case for TFIDF since matching based on these words should produce good results. However in the ``Hard'' case:

\begin{quotation}
\textbf{Hard Citance}: ``A lot of work has been done in English for the purpose of anaphora resolution and various algorithms have been devised for this purpose (Aone and Bennette, 1996; Brenan , Friedman and Pollard, 1987; Ge, Hale and Charniak, 1998; Grosz, Aravind and Weinstein, 1995; McCarthy and Lehnert, 1995; Lappins and Leass, 1994; \textit{Mitkov, 1998}; Soon, Ng and Lim, 1999).''

\textbf{Reference Span}: ``We have described a robust, knowledge-poor approach to pronoun resolution which operates on texts preprocessed by a part-of-speech tagger.''
\end{quotation}

We can see that there is no overlap of salient words, between the citance and text span. Not only is the citance somewhat vague, but any semantic overlap is not exact. For instance, ``anaphora resolution'' and ``pronoun resolution'' refer to the same concept but do not match lexically.

\subsection{Frequency of Section Titles}

We analyzed the frequency of section titles for the chosen reference sentences. Our analysis only excludes document P05-1053 from consideration -- the document whose answer key was withheld. For each cited reference sentence, we looked at the title of the section in which it appears. The titles that appeared with the greatest frequency can be seen in Table~\ref{tab:sectionfreq}. To extract these section titles we looked at the parent nodes of sentences within the XML document. The ``title'' and ``abstract'' sections are special since they refer to parent nodes of type other than SECTION. Due to OCR noise, a few section names were wrong. We manually corrected these section names. Then, we performed a slight normalization by removing any `s' character that appeared in the end of a name. These results clearly show sentences that are cited are not uniformly distributed within a document.

\begin{table}
	\centering
	\begin{tabular}{l|r}
		Title & Relevant \% \\
		\hline \hline
		Introduction & 26.50\% \\
		Abstract & 5.76\% \\
		Title & 5.47\% \\
		Conclusion & 4.58\% \\
		Evaluation & 4.43\%  \\
		The approach & 4.28\% \\
        Previous Work & 3.84\% \\
		Potential for improvement & 3.55\%  \\
        Discussion & 2.21\% \\
        Summary & 2.07\%
	\end{tabular}
	\vspace{0.5em}
	\caption{Sections are ordered from most to least frequently relevant. Considers all documents except P05-1053.}
	\label{tab:sectionfreq}
\end{table}

\subsection{Preprocessing}
Before we use the dataset in our system we preprocessed the dataset to reduce the number of errors. The dataset has lots of errors due to the use of OCR techniques. Broken words, non-ascii characters, and formating problems in XML files are some examples of these problems. We performed the following preprocessing steps to reduce noise in the dataset. First, we manually went over the citances and reference sentences fixing broken words (those separated by hyphen, space, or some non-ascii characters). We automatically removed all non-ascii characters from citance and reference text. Finally, we manually fixed some misformatted XML files that were missing closing tags.

\begin{table}[b]
\centering
\caption{Abbreviations.}
\label{tab:legend}
\footnotesize
\renewcommand\arraystretch{1.2}
\begin{tabular}{|c|p{0.3\textwidth}|}
\hline
nltk\_tok & Uses tokenizer from NLTK \\
\hline
sk\_tok   & Uses tokenizer from Scikit Learn \\
\hline
nltk\_stop & Uses stop words from NLTK \\
\hline
sk\_stop & Uses stop words from Scikit Learn \\
\hline
wn\_ref & WordNet applied only to reference sentences \\
\hline
wn\_cit & WordNet applied only to citances \\
\hline
wn\_both & WordNet applied to all sentences \\
\hline
st & Uses stemming during preprocessing \\
\hline
$(l,u)$ & Only considers sentences where $l~\le~\#tokens~\le~u$ \\
\hline
\end{tabular}
\end{table}

\section{TFIDF Approach}

\begin{table*}[t]
\centering
\subfloat[Results on \textbf{development} set.\label{tab:dev-twicewn-tfidf}]{
\begin{tabular}{l|l|r|r|r}

  id & Config. & R@3 & P@3 & $F_1$ \\ \hline \hline

  1 & tfidf+nltk\_stop+nltk\_tok+ref\_wn+(8,70)  & 24.55\% & 12.33\% & 16.41\% \\  

  2 & tfidf+sk\_stop+nltk\_tok+(8,70) & 23.94\% & 12.02\% & 16.01\% \\

  3 & tfidf+sk\_stop+sk\_tok+ref\_wn+(8,70)  & 23.64\% & 11.87\% & 15.81\% \\

  4 & tfidf+sk\_stop+sk\_tok+cit\_wn+(8,70)  & 23.64\% & 11.87\% & 15.81\% \\

  5 & tfidf+nltk\_stop+nltk\_tok+cit\_wn+(8,70)  & 23.33\% & 11.72\% & 15.60\% \\   

  6 & tfidf+sk\_stop+nltk\_tok+st+ref\_wn+(8,70) & 23.03\% & 11.57\% & 15.40\% \\

  7 & tfidf+nltk\_stop+nltk\_tok+(8,70) & 22.73\% & 11.42\% & 15.20\% \\

  8 & tfidf+nltk\_stop+nltk\_tok+st+ref\_wn+(8,70) & 22.12\% & 11.11\% & 14.79\% \\ 

  9 & tfidf+nltk\_stop+nltk\_tok+st+both\_wn+(8,70)  & 21.82\% & 10.96\% & 14.59\% \\  

\end{tabular}
}
\\
\subfloat[Results on \textbf{test} set.\label{tab:test-twicewn-tfidf}]{
\begin{tabular}{l|l|r|r|r}

  id & Config. & R@3 & P@3 & $F_1$ \\ \hline \hline

  1 & tfidf+nltk\_stop+nltk\_tok+cit\_wn+(8,70)  & 22.50\%  & 10.29\% & 14.11\% \\

  2 & tfidf+nltk\_stop+nltk\_tok+ref\_wn+(8,70)  & 22.29\% & 10.19\% & 13.99\% \\    

  3 & tfidf+nltk\_stop+nltk\_tok+st+(15,70) & 21.67\% & 9.90\% & 13.59\% \\

  4 & tfidf+nltk\_stop+nltk\_tok+cit\_wn+(15,70) & 21.46\% & 9.81\%  & 13.46\% \\

  5 & tfidf+nltk\_stop+nltk\_tok+st+cit\_wn+(15,70)  & 21.25\% & 9.71\%  & 13.33\% \\  

  6 & tfidf+nltk\_stop+nltk\_tok+st+ref\_wn+(8,70) & 21.04\% & 9.62\%  & 13.20\% \\

  7 & tfidf+nltk\_stop+nltk\_tok+ref\_wn+(15,70) & 20.83\% & 9.52\%  & 13.07\% \\

  8 & tfidf+sk\_stop+sk\_tok+cit\_wn+(15,70) & 20.62\% & 9.43\%  & 12.94\% \\  

  9 & tfidf+nltk\_stop+nltk\_tok+(15,70)  & 20.42\% & 9.33\%  & 12.81\% \\

\end{tabular}
}
\caption{Performance of TFIDF with improvements to WordNet.}
\label{tab:double-table-tfidf}
\end{table*}

Our best performing system for the CL-SciSumm 2016 task was based on TFIDF. 
It achieved  13.68\% $F_1$-score on the test set for Task 1a. Our approach compares the TFIDF vectors of the citance and the sentences in the reference document.
Each reference sentence is assigned a score according to the cosine similarity between itself and the citance. There were several variations studied to improve our TFIDF system. Table~\ref{tab:legend} contains the abbreviations we use when discussing a particular configuration.

Stopwords were removed for all configurations. These words serve mainly to add noise, so their removal helps improve performance. There are two lists of stopwords used: one from sklearn (sk\_stop) and one from NLTK (nltk\_stop).

To remove the effect of using words in their different forms we used stemming (st) to reduce words to their root form. For this purpose, we use the Snowball Stemmer, provided by the NLTK package~\cite{DBLP:journals/nle/Xue11}.

WordNet has been utilized to expand the semantics of the sentence. We obtain the lemmas from the synsets of each word in the sentence. We use the Lesk algorithm, provided through the NLTK package~\cite{DBLP:journals/nle/Xue11}, to perform wordsense disambiguation. This is a necessary step before obtaining the synset of a word from WordNet. Each synset is a collection of lemmas. The lemmas that constitute each synset are added to the word vector of the sentence; this augmented vector is used when calculating the cosine similarity instead of the original vector. 
We consider three different methods of using WordNet: ref\_wn, cit\_wn, and both\_wn, which will be explained in Subsection~\ref{ssec:improvetfidf}.

Our first implementation of WordNet expansion increased coverage at the cost of performance. With the proper adjustments, we were able to improve performance as well. This is another example of how the details and tuning of the implementation are critical in dealing with short text.
Our new implementation takes care to only add a word once, even though it may appear in the synsets of multiple words of the sentence. Details are found in Subsection~\ref{ssec:improvetfidf}.

In the shared task, filtering candidate sentences by length improved the system's performance substantially. Short sentences are unlikely to be chosen; they are often too brief to encapsulate a concept completely.
Longer sentences are usually artifacts from the PDF to text conversion (for instance, a table transformed into a sentence).
We eliminate from consideration all sentences outside a certain range of number of words. In our preliminary experiments, we found two promising lower bounds on the number of words: 8 and 15. The only upper bound we consider is 70, which also reduces computation time since longer sentences take longer to score. Each range appears in the tables as an ordered pair $(min,max)$; e.g.\ 8 to 70 words it would appear as $(8,70)$. This process eliminates some of the sentences our system is supposed to retrieve so our maximum attainable $F_1$-score is lowered.

\begin{table*}
  \centering
  \begin{tabular}{l|p{0.8\textwidth}}
    Topic Number & Top Words \\
    \hline \hline
    1 & word, featur, relat, our, set, model, figur, tabl, annot, noun, corpus, if, rule, system, text, structur \\
    \hline
    5 & text, essay, featur, score, system, corpus, data, our, set, perform, train, word, model, method, sentenc \\
    \hline
    10 & question, answer, model, type, system, semant, questions, qa, pattern, retriev, word, user, base \\
    \hline
    28 & rule, phrase, german, grammar, system, word, our, verb, sentenc, al, translat, lexic, text, annot 
  \end{tabular}
  \caption{Top words for topics in LDA-4.}
	\label{tab:words}
\end{table*}

\subsection{Improvements on TFIDF} 
\label{ssec:improvetfidf}

The main drawback of the TFIDF method is its inability to handle situations where there is no overlap between citance and reference sentence. Thus, we decided to work on improving the WordNet expansion. In Table~\ref{tab:double-table-tfidf} we can see the performance of various different configurations, some which improve upon our previously best system. 

The first improvement was to make sure the synsets do not flood the sentence with additional terms. Instead of adding all synsets to the sentence, we only added unique terms found in those synsets. Thus, if a term appeared in multiple synsets of words in the sentence it would still only contribute once to the modified sentence.

While running the experiments, the WordNet preprocessing was only applied to the citances instead of both citances and reference sentences by accident. This increased our performance to \textbf{14.11\%} (first entry on Table~\ref{tab:test-twicewn-tfidf}). To further investigate this, we also ran the WordNet expansion on only the reference sentences. This led to another subtlety being discovered, but before we can elaborate we must explain how WordNet expansion is performed.

Conceptually, the goal of the WordNet preprocessing stage is to increase the overlap between the words that appear in the citance and those that appear in the reference sentences. By including synonyms, a sentence has a greater chance to match with the citance. The intended effect was for the citances and reference sentences to meet in the middle.

The steps taken in WordNet expansion are as follows: each sentence is tokenized into single word tokens; we search for the synsets of each token in WordNet; if a synset is found, then the lemmas that constitute that synset are added to the sentence. The small subtlety referred to before is the duplication of original tokens: if a synset is found, it must contain the original token, so the original token gets added once more to the sentence. This adds more weight to the original tokens. Before the discovery of one-sided WordNet expansion, this was a key factor in our TFIDF results.

In actuality, adding synonyms to all reference sentences was a step too far. We believe that the addition of WordNet synsets to both the reference sentences and citances only served to add more noise. Due to the number of reference sentences, these additional synsets impacted the TFIDF values derived. However, if we only apply this transformation to the citances, the impact on the TFIDF values is minimal.

We now had to experiment with applying WordNet asymmetrically: adding synsets to citances only (cit\_wn) and adding synsets to reference sentences only (ref\_wn). In addition, we ran experiments to test the effect of duplicating original tokens. This would still serve the purpose of increasing overlap, but lessen the noise we introduced as a result. We can see the difference in performance in Table~\ref{tab:dev-twicewn-tfidf} and Table~\ref{tab:test-twicewn-tfidf}. In the development set, applying WordNet only to the reference sentences with duplication performed the best with an $F_1$-score of 16.41\%. For the test set, WordNet applied to only the citances performs the best with 14.12\%.  Regardless, our experiments indicate one-sided WordNet leads to better results. In the following sections, experiments only consider one-sided WordNet use.

\begin{figure}
	\centering
	\includegraphics[width=0.4\textwidth]{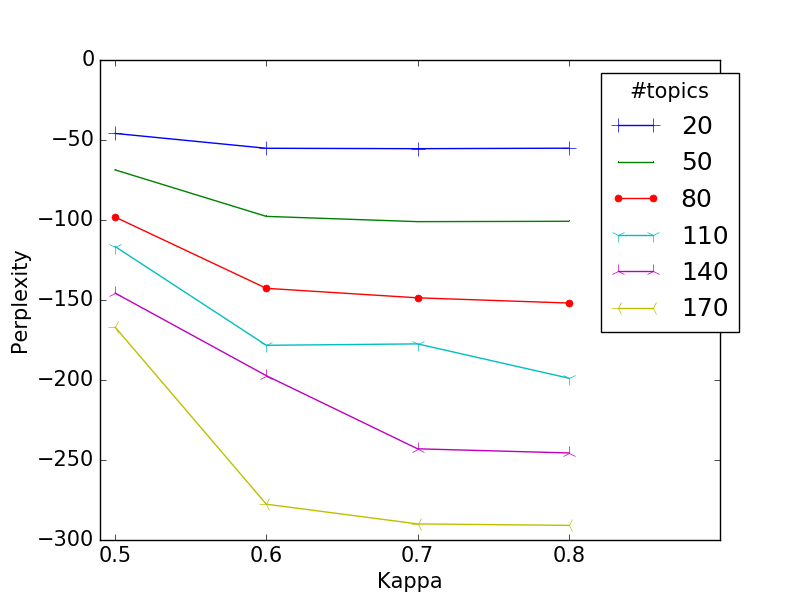}
	\caption{Perplexity values obtained by varying $\kappa$. Higher values are better.}
	\label{fig:perplexity}
\end{figure}

\begin{table*}
\centering
\subfloat[LDA topic models.\label{tab:ldaresults}]{
	\centering
	\begin{tabular}{l|r|r|r|r|r|r}
    id & \#topics & min\_df & max\_df & $\kappa$ & $\tau_0$ & $F_1$ \\
		\hline \hline
        1 & 95 & 10 & 0.99 & 0.5 & 768 & 7.90\% \\
        2 & 20 & 40 & 0.93 & 0.7 & 512 & 7.49\% \\
        3 &110 & 10 & 0.99 & 0.8 & 512 & 7.29\% \\
        4 &140 & 40 & 0.93 & 0.7 & 1   & 6.88\% \\
        5 & 20 & 40 & 0.87 & 0.7 & 256 & 6.88\%
	\end{tabular}
}
\\
\subfloat[Word Embeddings.\label{tab:weresults}]{
\centering
\begin{tabular}{l|l|r|r|r}

  id & Config. & R@3 & P@3 & $F_1$ \\ \hline \hline

  1 & WE-1+nltk\_stop+sk\_tok+st+cit\_wn+(8,70)  & 20.60\% & 10.35\% & 13.78\% \\  

  2 & WE-2+nltk\_stop+sk\_tok+st+cit\_wn+(8,70)  & 20.30\% & 10.20\% & 13.58\% \\  

  3 & WE-1+nltk\_stop+sk\_tok+st+ref\_wn+(8,70)  & 19.40\% & 9.74\% & 12.97\% \\  

  4 & WE-2+nltk\_stop+sk\_tok+st+ref\_wn+(8,70)  & 19.10\% & 9.59\% & 12.77\% \\  

  5 & WE-2+nltk\_stop+sk\_tok+st+(8,70)  & 18.18\% & 9.13\% & 12.16\% \\  

  6 & WE-1+nltk\_stop+sk\_tok+st+(8,70)  & 17.88\% & 8.98\% & 11.96\% \\  

\end{tabular}
}
\caption{Results on \textbf{development} set without TFIDF.}
\label{tab:notfidf}
\end{table*}

\section{Topic Modeling}

To overcome the limitations of using a single sentence we constructed topic models to better capture the semantic information of a citance. Using Latent Dirichlet Allocation (LDA), we created various topic models for the computational linguistics domain.

\subsection{Corpus Creation}
First, we gathered a set of 34273 documents from the ACL Anthology\footnote{\url{https://aclweb.org/anthology/}} website. This set is comprised of all PDFs available to download. The next step was to convert the PDFs to text. Unfortunately, we ran into the same problem as the organizers of the shared task: the conversion from PDF to text left a lot to be desired. Additionally, some PDFs used an internal encoding thus resulting in an undecipherable conversion. Instead of trying to fix these documents, we decided to select a subset that seemed sufficiently error-free. Since poorly converted documents contain more symbols than usual, we chose to cluster the documents according to character frequencies. Using K-means, we clustered the documents into twelve different clusters.

After clustering, we manually selected the clusters that seemed to have articles with acceptable noise. Interestingly, tables of content are part of the PDFs available for download from the anthology. Since these documents contain more formatting than text, they ended up clustered together. We chose to disregard these clusters as well. In total, 26686 documents remained as part of our ``cleaned corpus''.

\subsection{Latent Dirichlet Allocation}
\label{ssec:lda}

LDA can be seen as a probabilistic factorization method that splits a term-document matrix into term-topic and topic-document matrices. The main advantage of LDA is its soft clustering: a single document can be part of many topics to varying degree.

We are interested in the resulting term-topic matrix that is derived from the corpus. With this matrix, we can convert terms into topic vectors where each dimension represents the term's extent of membership. These topic vectors provides new opportunities for achieving overlap between citances and reference sentences, thus allowing us to score sentences that would have a cosine similarity of zero between TFIDF vectors.

Similar to K-means, we must choose the number of topics beforehand. Since we are using online LDA \cite{DBLP:conf/nips/HoffmanBB10} there are a few additional parameters, specifically: $\kappa$, a parameter to adjust the learning rate for online LDA; $\tau_0$, a parameter to slow down the learning for the first few iterations. The ranges for each parameter are [0.5, 0.9] in increments of 0.1 for $\kappa$ and 1, 256, 512, 768 for $\tau_0$.

We also experimented with different parameters for the vocabulary. The minimum number of documents in which a word had to appear was an absolute number of documents (min\_df): 10 or 40. The maximum number of documents in which a word could appear was a percentage of the total corpus (max\_df): 0.87, 0.93, 0.99.

One way to evaluate the suitability of a learned topic model is through a measure known as perplexity \cite{DBLP:conf/nips/HoffmanBB10}. 
Since LDA learns a distribution for topics and terms, we can calculate the probability of any document according to this distribution.
Given an unseen collection of documents taken from the same domain, we calculate the probability of this collection according to the topic model. We expect a good topic model to be less ``surprised'' at these documents if they are a representative sample of the domain. In Figure~\ref{fig:perplexity}, we graph the perplexity of our topic models when judged on the reference documents of the training and development set of the CL-SciSumm dataset.

Unfortunately, the implementation we used does not normalize these values, which means we cannot use the perplexity for comparing two models that have a different number of topics. Keep in mind the numbers in Figure~\ref{fig:perplexity} do not reflect the perplexity directly. Perplexity is still useful for evaluating our choice of $\kappa$ and $\tau_0$. We omit plotting the perplexity for different $\tau_0$ values since, with regards to perplexity, models with $\tau_0 > 1$ always underperformed. Figure~\ref{fig:perplexity} makes a strong case for the choice of $\kappa=0.5$. However, our experiments demonstrate that, for ranking, higher $\kappa$ and higher $\tau_0$ can be advantageous.

In order to compare these models across different number of topics we evaluated their performance at Task 1a. The results of these runs can be seen in Table~\ref{tab:ldaresults}. Sentences were first converted to LDA topic vectors then ranked by their cosine similarity to the citance (also a topic vector). The performance of this method is worse than all TFIDF configurations, regardless of which LDA model is chosen. We merely use these results to compare the different models. Nevertheless, the topics learned by LDA were not immediately evident -- suggesting there is room for improvement in the choice of parameters. Table~\ref{tab:words} has a selection of the most interpretable topics for one of the models.

\section{Word Embeddings}
\label{sec:wordembed}

Another way to augment the semantic information of the sentences is through word embeddings. The idea behind word embeddings is to assign each word a vector of real numbers. These vectors are chosen such that if two words a similar, their vectors should be similar as well. We learn these word embeddings in the same manner as word2vec \cite{word2vec}.

We use DMTK \cite{DMTK} to learn our embeddings. DMTK provides a distributed implementation of word2vec.
We trained two separate embeddings: WE-1 and WE-2. We only explored two different parameter settings. Both embeddings consist of a 200 dimensional vector space. Training was slightly more intensive for WE-1, which ran for 15 epochs sampling 5 negative examples. The second embedding, WE-2, ran for 13 epochs sampling only 4 negative examples. The minimum count for words in the vocabulary was also different: WE-1 required words to appear 40 times whereas WE-2 required words to appear 60 (thus, resulting in a smaller vocabulary).

To obtain similarity scores, we use the \emph{Word Mover's Distance} \cite{wmdistance}. Thus, instead of measuring the similarity to an averaged vector for each sentence, we consider the vector of each word separately. In summary, given two sentences, each composed of words which are represented as vectors, we want to move the vectors of one sentence atop those of the other by moving each vector as little as possible. The results obtained can be found in Table~\ref{tab:weresults}.

Word embeddings outperformed topic models on the development set; while the highest scoring topic model achieved $7.90\%$ $F_1$-score on the development set, the highest scoring word embedding achieved $13.77\%$.

\begin{figure}
	\centering
	\includegraphics[width=0.5\textwidth]{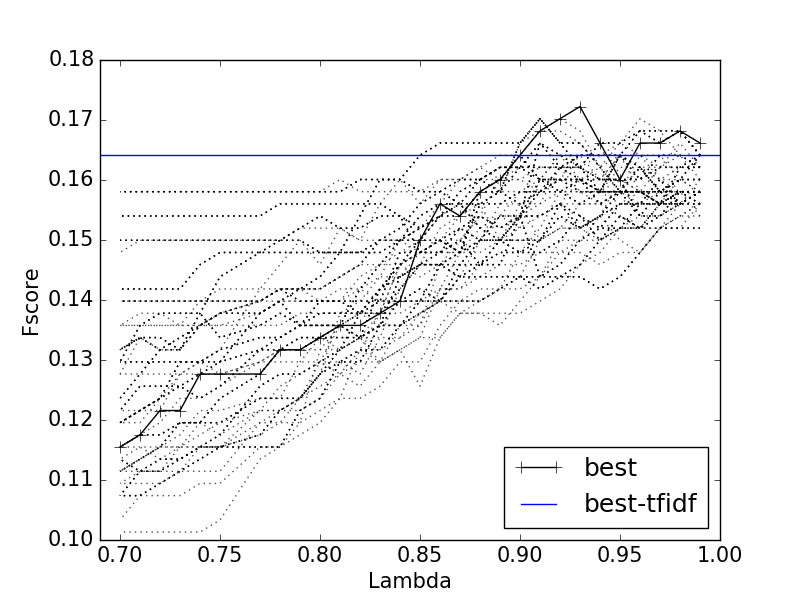}
  \caption{$F_1$-score for TFIDF + LDA configurations on \textbf{development} set with different  $\lambda$ values.}
	\label{fig:lambda}
\end{figure}

\begin{table}
	\centering
	\begin{tabular}{l|r|r|r|r}
    Config. & $\lambda$ range &  R@3  & P@3 & $F_1$    \\
		\hline \hline
        T1 + WE-1 & $0.70-0.70$ & 25.76\% & 12.94\% & 17.22\% \\
        T1 + WE-1 & $0.71-0.71$ & 25.45\% & 12.79\% & 17.02\% \\
        T1 + WE-1 & $0.93-0.94$ & 25.15\% & 12.63\% & 16.82\% \\
        T1 + WE-1 & $0.75-0.82$ & 25.15\% & 12.63\% & 16.82\% \\
        T1 + WE-1 & $0.95-0.99$ & 24.85\% & 12.48\% & 16.62\% \\
        \hline
        T1 + WE-2 & $0.93-0.94$ & 25.15\% & 12.63\% & 16.82\% \\
        T1 + WE-2 & $0.76-0.82$ & 25.15\% & 12.63\% & 16.82\% \\
        T1 + WE-2 & $0.70-0.70$ & 25.15\% & 12.63\% & 16.82\% \\
        T1 + WE-2 & $0.95-0.99$ & 24.85\% & 12.48\% & 16.62\% \\
        T1 + WE-2 & $0.83-0.92$ & 24.85\% & 12.48\% & 16.62\% \\
	\end{tabular}
	\vspace{0.5em}
    \caption{Results for TFIDF + WE configurations on \textbf{development} set.}
	\label{tab:welambda}
\end{table}

\section{Tradeoff Parameterization}
\label{sec:tradeoff}

In order to combine the TFIDF systems with LDA or Word Embedding systems, we introduce a parameter to vary the importance of the added similarity compared to the TFIDF similarity: $\lambda$. The equation for the new scores is thus: 
\begin{equation}
\lambda \cdot TFIDF \enskip + \enskip (1-\lambda) \cdot other \label{trade}
\end{equation}
where $other$ stands for either LDA or WE.

Each sentence is scored by each system separately. These two values (the TFIDF similarity and the other system's similarity) are combined through Equation~\ref{trade}. The sentences are then ranked according to these adjusted values.

We evaluated this method by taking the 10 best performing systems on the development set for both TFIDF and LDA. Each combination of a TFIDF system with an LDA system was tested. We test these hybrid systems with values of $\lambda$ between $[0.7, 0.99]$ in $0.01$ increments. There were only 6 different configurations for word embeddings so we used all of them with the same values for $\lambda$.

After obtaining the scores for the development set, we chose the 100 best systems to run on the test set (for LDA only). Systems consist of a choice of TFIDF system, a choice of LDA system, and a value for $\lambda$. The five highest scoring systems are shown in Table~\ref{tab:lambda-test}.

We can see that a particular topic model dominated. The LDA model that best complemented any TFIDF system was only the fourth best LDA system on the development set. There were multiple combinations with the same $F_1$-score, so we had to choose which to display in Tabel~\ref{tab:lambda-test}. This obscures the results since other models attain $F_1$-score as high as $14.64\%$. In particular, the second best performing topic model in this experiment was an 80 topic model that is not in Table~\ref{tab:ldaresults}. The following best topic model had 50 topics.

Interestingly, a TFIDF system coupled with word embeddings performs incredibly well on the development set as can be seen in Table~\ref{tab:welambda} (values similar to LDA if we look at Fig~\ref{fig:lambda}). However, once we move to the test set, all improvements become meager. It is possible that word embeddings are more sensitive to the choice of $\lambda$.

Although we do not provide the numbers, if we analyze the distribution of scores given by word embeddings, we find that the distribution is much flatter. TFIDF scores drop rapidly; for some citances most sentences are scored with a zero. LDA improves upon that by having less zero-score sentences, but the scores still decay until they reach zero. Word embeddings, however, seem to grant a minimum score to all sentences (most scores are greater than 0.4). Furthermore, there is very little variability from lowest to highest score. This is further evidenced by the wide range of $\lambda$ values that yield good performance on Table~\ref{tab:welambda}. We conjecture the shape of these distributions may be responsible for the differences in performance.

\begin{table}
	\centering
	\begin{tabular}{l|r|r|r|r}
    Config.  & $\lambda$ & R@3 & P@3 & $F_1$ \\
		\hline \hline
        T1 + LDA-4 & 0.93 & 23.54\% & 10.76\% & 14.77\% \\
        T1 + LDA-4 & 0.95 & 23.33\% & 10.66\% & 14.64\% \\
        T1 + LDA-4 & 0.96 & 22.91\% & 10.47\% & 14.37\% \\
        T1 + LDA*  & 0.96 & 22.70\% & 10.38\% & 14.24\% \\
        T1 + LDA-4 & 0.91 & 22.29\% & 10.19\% & 13.98\% \\
		\hline
        T1 + WE-2 & 0.71 & 22.70\% & 10.38\% & 14.24\% \\
        T1 + WE-1/2 & 0.72 & 22.50\% & 10.28\% & 14.11\% \\
        T1 + WE-1/2 & 0.77 & 22.29\% & 10.19\% & 13.98\% \\
        T1 + WE-1  & 0.75 & 20.08\% & 10.09\% & 13.85\%
	\end{tabular}
	\vspace{0.5em}
    \caption{Results for tradeoff variations on \textbf{test} set. The model LDA* is not in Table~\ref{tab:ldaresults}. WE-1/2 means either version.}
	\label{tab:lambda-test}
\end{table}

\section{Statistical Analysis}
\label{sec:stats}

Although the $F_1$-scores have improved by augmenting the bare-bones TFIDF approach, we must still check whether this improvement is statistically significant. Since some of these systems have very similar $F_1$-scores, we cannot simply provide a 95\% confidence intervals for each $F_1$-score individually; we are forced to perform paired t-tests which mitigate the variance inherent in the data.

Given two systems, A and B, we resample with replacement 10000 times the dataset tested on and calculate the $F_1$-score for each new sample. By evaluating on the same sample, any variability due to the data (harder/easier citances, for instance) is ignored. Finally, these pairs of $F_1$-scores are then used to calculate a p-value for the paired t-test.

We calculate the significance of differences between the top entries for each category (TFIDF and two tradeoff variations) evaluated on the test set. For the best performing system (that is TFIDF + LDA at $F_1$-score of $14.77\%$) the difference is statistically significant from TFIDF + WE (at $14.24\%$) and TFIDF (at $14.11\%$) with p-values of $0.0015$ and $0.0003$, respectively. However, the difference between TFIDF + WE and TFIDF is not statistically significant (p-value of $0.4830$).

\section{Human Annotators}
\label{sec:human}

\begin{table}[b]
	\centering
	\caption{Human annotator comparison.}
	\label{tab:human}
	\begin{tabular}{l|r|r|r}
	Annotator & R@3 & P@3 & $F_1$ \\
		\hline \hline
		System ($\lambda =0.93$) & 25.60\% & 12.91\% & 17.17\% \\
		System ($\lambda =0.95$) & 24.40\% & 12.31\% & 16.37\% \\
		System ($\lambda =0.96$) & 23.80\% & 12.01\% & 15.97\% \\
        \hline
		Human 1 & 23.81\% &  3.21\% &  5.66\% \\
		Human 2 & 47.62\% &  7.84\% & 13.46\% \\
		Human 3 & 26.79\% & 29.22\% & 27.95\% \\
		Human 1 (title adj.) & 20.83\% & 22.15\% & 21.47\% \\
		Human 2 (title adj.) & 38.69\% & 13.00\% & 19.46\%
	\end{tabular}
\end{table}

In order to determine whether the performance of our system is much lower than what can be achieved, we ran an experiment with human annotators. Since human annotators require more time to perform the task, we had to truncate the test set to just three documents, chosen at random.

The subset used to evaluate the human annotators consists of three different articles from the test set: C00-2123, N06-2049, and J96-3004. The 20 citances that cite C00-2123 only select 24 distinct sentences from the reference article, which contains 203 sentences. Similarly, the 22 citances that cite N06-2049 select only 35 distinct reference sentences from 155 total. The last article, J96-3004, has 69 citances annotated that select 109 distinct reference sentences from the 471 sentences found in the article.

To avoid ``guiding'' the annotators to the correct answers, we provided minimal instructions. We explained the problem of matching citances to their relevant reference spans to each annotator. Since the objective was to compare to our system's performance, the annotators had at their disposal the XML files given to our system. Thus, the sentence boundaries are interpreted consistently by our system and the human annotators. We instructed them to choose one or many sentences, possibly including the title, as the reference span for a citance.

The performance of the human annotators and three of our best system configurations can be seen in Table~\ref{tab:human}. The raw score for two of the annotators had extremely low precision. Upon further analysis, we noticed outliers where more than ten different reference sentences had been chosen.

To provide a fairer assessment, the scores were adjusted for two of the annotators: if more than ten sentences were selected for a citance, we replace the sentences with simply the article title. We argue this is justified since any citance that requires that many sentences to be chosen is probably referencing the paper as a whole. After these adjustments, the score of the human annotators rose considerably.

\section{Discussion}
\label{sec:discussion}

The fact that WordNet went from decreasing the performance of our system to increasing its performance shows the level of detail required to tune a system for the task of reference span identification. The performance of our human annotators demonstrate the difficulty of this task -- it requires much more precision. Additionally, the human scores show there is room for improvement.

Task 1a can be framed as identifying semantically similar sentences. This perspective is best represented by the LDA systems of Section~\ref{ssec:lda} and the word embeddings of Section~\ref{sec:wordembed}. However, as can be seen by the results we obtained in Table~\ref{tab:notfidf}, relying solely on semantic similarity is not the best approach.

Methods such as TFIDF and sentence limiting do not attempt to solve Task 1a head-on. Through a narrowing of possibilities, these methods improve the odds of choosing the correct sentences. Only after sifting through the candidate sentences with these methods can topic modeling be of use. 

Combining TFIDF and LDA through a tradeoff parameter allowed us to test whether topic modeling does indeed improve our performance. Clearly, that is the case since our best performing system uses both TFIDF and LDA. The same experiment was performed with word embeddings, although the improvements were not as great.

Since word embeddings performed well alone but didn't provide much of a boost to TFIDF, it is possible the information captured by the embeddings overlaps with the information captured by TFIDF.

The question that remains is whether the topic modeling was done as best as it could be. The results in Section~\ref{sec:tradeoff} require further analysis. As we can see from Figure~\ref{fig:lambda}, very few combinations provide a net-gain in performance.
Likewise, it is possible that further tuning of word embedding parameters could improve our performance.

\section{Conclusion}

During the BIRNDL shared task, we were surprised by the result of our TFIDF system, which achieved an $F_1$-score of 13.65\%. More complex systems did not obtain a higher $F_1$-score. In this paper, we show it is possible to improve our TFIDF system with additional semantic information.

Through the use of WordNet we achieve an $F_1$-score of 14.11\%. Word embeddings increase this $F_1$-score to 14.24\%. If we employ LDA topic models instead of word embeddings, our system attains its best performance: an $F_1$-score of 14.77\%. This improvement is statistically significant.

Although these increases seem modest, the difficulty of the task should be taken into account. We performed an experiment with human annotators to assess what $F_1$-score would constitute a reasonable goal for our system. The best $F_1$-score obtained by a human was 27.95\%. This leads us to believe there is still room for improvement on this task.

In the future, the study of overlap between TFIDF and word embeddings could provide a better understanding of the limits of this task. Finally, we also propose the simultaneous combination of LDA topic models and word embeddings.

\section{Acknowledgments}

We would like to thank Avisha Das, Arthur Dunbar, and Mahsa Shafaei for providing the annotations.

The authors acknowledge the use of the Maxwell Cluster and the advanced support from the Center of Advanced Computing and Data Systems at the University of Houston to carry out the research presented here.

\bibliographystyle{spmpsci}      
\bibliography{ref}   

\end{document}